# Computational, Data-Driven, and Physics-Informed Machine Learning Approaches for Microstructure Modeling in Metal Additive Manufacturing

D. Patel[a,b], R. Sharma[a,b], Y.B. Guo[a,b]

[a] Dept. of Mechanical and Aerospace Engineering, Rutgers University-New Brunswick, Piscataway, NJ 08854, USA
[b] New Jersey Advanced Manufacturing Initiative, Rutgers University-New Brunswick, Piscataway, NJ 08854, USA

*Corresponding author: yuebin.guo@rutgers.edu (Y.B. Guo)

**ABSTRACT**

*Metal additive manufacturing (AM) enables unprecedented design freedom and the production of customized, complex components. However, the rapid melting and solidification dynamics inherent to metal AM processes generate heterogeneous, non-equilibrium microstructures that significantly impact mechanical properties and subsequent functionality. Predicting microstructure and its evolution across spatial and temporal scales remains a central challenge for process optimization and defect mitigation. While conventional experimental techniques and physics-based simulations provide a physical foundation and valuable insights, they face critical limitations such as high computational cost, limited scalability, and difficulty bridging across scales. In contrast, data-driven machine learning (ML) offers an alternative prediction approach and powerful pattern recognition but often operate as "black-box", lacking generalizability and physical consistency, particularly in data-scarce scenarios. To overcome these limitations, physics-informed machine learning (PIML), including physics-informed neural networks (PINNs), has emerged as a promising paradigm by embedding governing physical laws into neural network architectures, thereby enhancing accuracy, transparency, data efficiency, and extrapolation capabilities. This work presents a comprehensive evaluation of modeling strategies for microstructure prediction in metal AM. The strengths and limitations of experimental, computational, and data-driven methods are analyzed in depth, and highlight recent advances in hybrid PIML frameworks that integrate physical knowledge with ML. Key challenges, such as data scarcity, multi-scale coupling, and uncertainty quantification, are discussed alongside future directions. Ultimately, this assessment underscores the importance of PIML-based hybrid approaches in enabling predictive, scalable, and physically consistent microstructure modeling for site-specific, microstructure-aware process control and the reliable production of high-performance AM components.*

## 1. INTRODUCTION

### 1.1 Microstructures in Metal Additive Manufacturing

Metal AM has emerged as a transformative technology for producing complex geometry metal components, multi-material components, and functionally graded materials [1], across various industries, including aerospace, biomedical, tooling, automotive, and energy [2]. By overcoming the design constraints of traditional manufacturing methods, such as machining and forming, metal

AM has gained significant popularity in the past decade. Applications range from lightweight, topology-optimized aerospace components and patient-specific implants to complex heat exchangers and fuel nozzles [2,3].

According to the ASTM F2792 standard and the more recent ASTM F52900, metal AM processes are broadly categorized into Powder Bed Fusion (PBF) and Directed Energy Deposition (DED) [2]. PBF selectively melts successive layers of powder using a high-energy laser or electron beam, while DED involves the simultaneous deposition and melting of powder or wire feedstock using a focused laser source. Both processes are characterized by rapid melting and solidification with high thermal gradients and cooling rates, leading to complex microstructure (including defects) formation and evolution.

From a metallurgical standpoint, the rapid solidification inherent to melt pool results in complex thermal histories and non-equilibrium conditions that strongly influence the microstructure and properties of the final part. The steep thermal gradients and cooling rates, often exceeding $10^4$–$10^6$ K/s, can give rise to microstructural features such as hierarchical solidification structures, anisotropic grain growth, and metastable phase formations [4]. For instance, many alloys exhibit a tendency toward elongated columnar grain growth along the build direction, which causes anisotropic mechanical behavior. Such characteristics are often undesirable in applications requiring isotropic properties, particularly in critical aerospace and biomedical components [5]. Furthermore, the nature of cyclic and localized heat input during AM introduces complex melt pool dynamics that may lead to process-induced defects, including porosity (e.g., keyhole or lack-of-fusion (LOF) pores), cracking, and residual stresses. These microstructural heterogeneities and defects not only compromise mechanical performance but also impose significant challenges for part certification and quality assurance [6]. These issues can be mitigated through tailored materials, process optimization, real-time monitoring, interlayer deformation, and both intrinsic and post-process heat treatments. Achieving "first-time-right" builds remains a critical barrier to the broader industrial adoption of metal AM, especially in safety-critical applications.

Addressing these challenges necessitates a comprehensive understanding of the complex microstructures in metal AM. The microstructure formation process is highly nonlinear and governed by numerous interdependent process parameters such as laser power, scanning speed, hatch spacing, and layer thickness [7]. Extensive research efforts have been dedicated to elucidating microstructure formation through a combination of experimental characterization, computational modeling, and data-driven approaches. Experimental techniques remain as a foundation for quantifying microstructural features, such as grain morphology, crystallographic texture, phase distribution, and defect types, but the extreme processing conditions of AM processes often hinder direct in-situ monitoring. Therefore, computational modeling has become a critical tool for probing the thermal, mechanical, and metallurgical phenomena during AM [8]. Finite Element Analysis (FEA) is commonly employed to predict temperature profiles and residual stresses [9], while mesoscale techniques such as Phase-Field (PF), Cellular Automata (CA) and Kinematic Monte Carlo (KMC) modeling are used to simulate grain growth and solidification dynamics [10]. In recent years, data-driven methods, particularly ML approaches such as PINNs, have emerged as powerful tools for microstructure modeling and optimization [11]. In summary, unlocking the full potential of metal AM hinges on the integration of experimental insights, multi-physics simulations, and data-driven models to establish predictive approaches. Such integrated approaches are key to predicting microstructure formation and characteristics, tailoring microstructures, minimizing defects, and ensuring the repeatable production of high-performance, application-specific components.

## 1.2 Data-Driven Approach for Microstructure Prediction

In-process sensing and monitoring are essential for enabling real-time quality control and improving microstructural predictability in metal AM. Various sensor modalities, including optical (high-speed cameras, vision systems), thermal (infrared (IR) cameras, pyrometers), acoustic emission (AE) sensors, and spectral devices, are widely used to capture key process signatures such as melt pool geometry [3], temperature distribution, plume dynamics, spatter, and surface morphology [12].These measured signals provide critical insight into the transient and localized conditions during the build, which are directly linked to microstructural outcomes such as grain size, orientation, porosity, and residual stresses. As such, in-situ monitoring offers a non-destructive means of observing process stability, detecting defects, and identifying deviations in real time, forming the foundation for data-driven process control strategies [8].

Recent research emphasizes the integration of large sensor data with computational models to enhance predictive capabilities. Data-driven approaches using ML and deep learning (DL) have been successful in analyzing sensor signals to predict melt pool behavior, detect defects, and forecast microstructural features[14]. Hybrid approaches, enhance robustness and generalizability in data-scarce conditions. For instance, Bevans et al. [13] developed a digital twin framework for Inconel 718 by integrating IR camera-based thermal histories, physics-based thermal models, and ML algorithms to predict melt pool depth and grain size. This illustrates how combining sensor data with physical insights improves microstructure prediction. However, challenges remain in handling data uncertainty, ensuring cross-platform applicability, and achieving real-time responsiveness [14]. Despite this, the continued advancement of in-situ sensing technologies and modeling techniques holds great promise for realizing intelligent and adaptive AM processes with tailored microstructures and improved part quality.

The future of metal AM is moving toward intelligent and autonomous systems, powered by advanced in-situ monitoring, sensor fusion, and real-time control. Recent reviews have highlighted the importance of combining sensing technologies with automated feedback to improve the quality and consistency of AM builds. For instance, Mu et al. [15] proposed a digital twin framework for metal-DED that helps optimize the process through virtual simulations. Similarly, Cai et al. [16] reviewed sensing and control in laser-based AM, showcasing progress in sensors, data collection, and control algorithms aimed at reducing defects and increasing reliability. Herzog et al. [17] also emphasized the role of ML in defect-detection and quality assurance through real-time monitoring. These advancements are leading toward AM systems that can automatically adjust laser power, scan speed, or material flow during printing, reducing defects and minimizing the need for post-processing. However, challenges remain, especially in ensuring sensor reliability, handling large volumes of data, and creating adaptive control strategies that work across different AM machines and materials.

Researchers are also exploring PIML, which combines physical laws with ML models. While traditional ML needs large, high-quality datasets and often lacks physical meaning, PIML improves accuracy and generalization by using built-in physical principles. Future research is looking at more advanced PIML methods, such as using physics in training (PIMT), model design (PIMA), components (PIMC), and output constraints (PIMO) [18]. Hybrid models that combine different types of data, like thermal images, sound, and high-speed videos, will further strengthen real-time prediction and control of microstructures [19,20]. In addition, multiscale PIML models that capture behavior from the micro to macro level are essential to fully understand and control the link between process, structure, and properties in metal AM [21]. These approaches aim to create smart, adaptive AM systems that are highly precise, efficient, and reliable [20].

## 1.3 Paper Structure

To provide an in-depth assessment of microstructure prediction in metal additive manufacturing, a structured literature review is conducted, focusing on ML approaches, with particular emphasis on PIML, including PINNs as a prominent subset. The review emphasizes the recent literature that combines experimental methods, data-driven modeling, and physics-based approaches, particularly in the context of microstructure evolution in LPBF and DED processes. Relevant articles were identified through targeted searches using Google Scholar and Scopus, with selection criteria based on publication date (primarily from 2018 to 2024), relevance to ML, PINN, and PIML, and a decade-long span for studies focusing on process parameter effects on microstructure. Additional references were incorporated through backward citation tracing of key articles. For document analysis and synthesis, NotebookLM was used to assist in extracting, organizing, and summarizing key insights from the selected literature. Additionally, ChatGPT and Gemini played a significant role in summarizing, analyzing, and structuring the extracted content. These AI tools facilitated the identification of key research themes, the rephrasing of complex sections, and the organization of ideas. Perplexity AI were also consulted for supplementary article suggestions and for validating exploratory topics, ensuring a comprehensive and up-to-date review of the field.

The structure of the article is organized as follows: Section 2 outlines the fundamentals of microstructure formation in metal AM, emphasizing the role of melt pool dynamics, thermal gradients, and solidification behavior. Section 3 presents conventional experimental methods used for microstructural analysis and model validation. Section 4 discusses physics-based modeling approaches that simulate microstructure evolution using governing physical principles. Section 5 reviews data-driven and hybrid strategies, highlighting the integration of physical knowledge into machine learning frameworks to enhance prediction accuracy and generalizability. Section 6 discusses current challenges in the field, such as data scarcity, generalizability, and physical consistency. Finally, Section 7 concludes with a summary and highlights future research directions toward integrating PIML for microstructure-aware AM process design.

## 2. COMPLEX MELT POOL PHENOMENA IN METAL AM

### 2.1 High Temperature Gradients and Cooling Rates

Metal AM processes, notably LPBF and DED, are characterized by extreme thermal conditions unlike conventional methods. These involve very high melt pool temperatures (up to 3000 °C) and steep thermal gradients (G), which, combined with the solidification rate (R), lead to exceptionally high cooling rates (G×R), potentially reaching $10^5$–$10^7$ K/s in LPBF [22] and $10^2$–$10^5$ K/s in DED [23]. These thermal conditions, heavily influenced by process parameters like laser power and scan speed [8,23], are critical determinants of the final microstructure. High cooling rates (G×R) generally promote microstructural refinement, reducing primary dendrite spacing and grain size, as faster cooling enhances nucleation and limits growth [23]. Studies have shown that lower heat input correlates with higher G and improved strength [24], while rapid cooling near melt pool boundaries fosters grain refinement [23]. Conversely, slower cooling, perhaps due to shorter inter-layer times or higher energy input, can result in coarser grains [25].

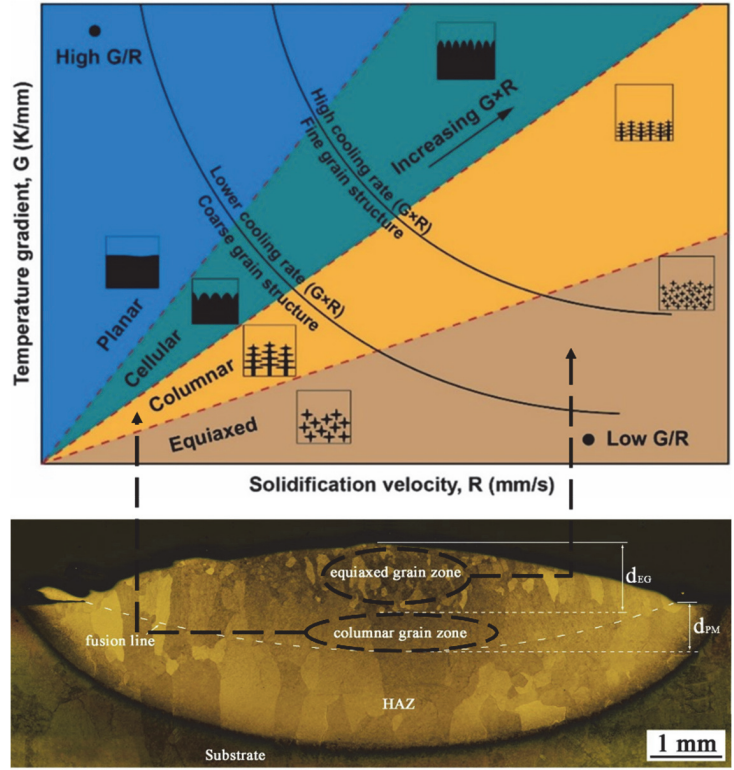

**FIG. 1:** Solidification map as a function of temperature gradient and solidification rate [27] (top), and micrograph showing the formation of equiaxed and columnar grains in a solidified melt pool [29] (bottom).

The morphology of the solidifying structure is primarily governed by the ratio G/R [26,27]. As depicted in solidification maps (Fig. 1, top), high G/R values favor directional solidification, often leading to columnar grains growing epitaxially layer upon layer, frequently aligned with the build direction and contributing to crystallographic texture. Lower G/R values tend to promote the formation of equiaxed grains. Transitions between these morphologies can occur within the melt pool itself due to local variations in G and R [28,29], resulting in mixed microstructures, as observed experimentally in alloys like Ti (Fig. 1, bottom). Furthermore, the rapid solidification inherent to AM often suppresses equilibrium phase transformations, leading to the formation of metastable phases, such as $\alpha'$-martensite in titanium alloys [4] or $\varepsilon$-/$\alpha'$-martensite in certain steels [30]. Repeated thermal cycling also induces intrinsic heat treatment (IHT) effects, driving solid-state transformations like precipitation or tempering, and can lead to elemental segregation, forming features like Laves phases in Inconel 718 [31] or influencing precipitation in AlSi10Mg [32].

While beneficial for achieving fine microstructures and potentially enhanced strength/hardness, these extreme thermal conditions also pose challenges [33]. Steep thermal gradients contribute to high residual stresses, potentially causing part distortion or cracking [8]. Improper energy input, either too low or too high, can respectively cause lack-of-fusion porosity or keyhole-induced porosity and overheating [34]. The resulting microstructural features, such as grain size, morphology, texture, phase distribution, and defects, collectively dictate the final mechanical performance, including strength, ductility, and anisotropy [8]. Therefore, precise control and understanding of the thermal phenomena during AM are essential for tailoring microstructures and optimizing component integrity and properties.

## 2.2 Melt Pool Dynamics

Melt pool dynamics play a central role in microstructure evolution during metal AM. High thermal gradients, rapid solidification, and Marangoni convection in the melt pool create complex transient conditions that influence grain morphology, phase formation, and defect generation. Key process parameters, such as laser power, scanning speed, and hatch spacing, directly affect melt pool geometry and thermal history, thus shaping the solidification path, as shown in Fig. 2. Solidification parameters, namely the temperature gradient (G) and solidification rate (R), jointly determine microstructural features via the G/R and GR values. Repeated thermal cycling in multi-layer builds introduces further complexity through remelting, heat treatment, and recrystallization. Computational modeling, supported by experimental methods like EBSD and synchrotron imaging, enables the prediction of melt pool behavior and microstructure. The following subsections explain the different phenomena in the melt pool and their influence on melt pool geometry.

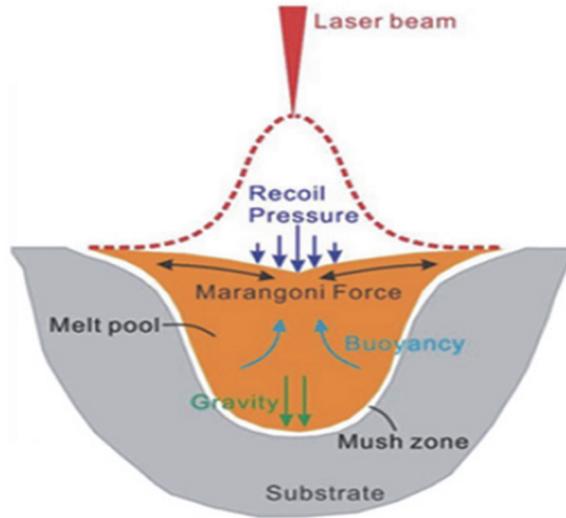

**FIG. 2**: Different multiphysics and forces acting on the melt pool [2].

**Marangoni convection:** Marangoni flow is a surface-tension-driven phenomenon that plays a significant role in metal AM. It arises from the presence of surface tension gradients along the free surface of the melt pool, which are typically induced by temperature variation [27,35–37]. In most metallic liquids, surface tension exhibits an inverse relationship with temperature, meaning that hotter regions possess lower surface tension compared to cooler regions [18,37]. This gradient in surface tension generates a tangential force that drives the fluid flow from areas of lower surface tension (hotter, often the center of the laser beam) towards areas of higher surface tension (cooler, often the edges of the melt pool) [36]. However, as highlighted by multiple studies, the presence of surface-active elements, such as sulfur in steels, can significantly alter this behavior, potentially reversing the flow direction to inward (from cooler edges to the hotter center) [3,27]. These Marangoni-induced flow patterns are fundamental in shaping the melt pool's size, depth, and overall dynamics, which directly dictates the conditions under which the material solidifies.

The profound influence of Marangoni convection extends directly to the resulting microstructure of the additively manufactured part. By dictating the fluid flow and heat distribution within the melt pool, Marangoni forces modify the crucial solidification parameters, specifically the thermal gradient (G), the solidification rate (R), and their ratio (G/R). These conditions determine the size and shape of the metal grains. Marangoni convection directly affects solidification rates and thermal gradients, which are critical in determining grain growth and crystallographic texture [3]. For instance, process-microstructure models have shown that convection affects grain structure by influencing nucleation events and the direction of grain growth. For instance, stronger Marangoni flow can enhance bulk nucleation and promote equiaxed grains [37], though its impact on grain morphology may be limited under some conditions [35]. Its influence is indirect, acting through melt pool shape and thermal gradients. The flow can also

Page **6** of 36

influence the overall grain alignment, called texture, and can even be strong enough to bend growing crystal structures called dendrites [33]. Hence, the Marangoni convection determines the melt pool behavior and consequently impacts the microstructure.

**Natural convection:** Natural convection in LPBF and DED arises from buoyancy forces due to temperature-induced density gradients in the melt pool [27,37], lighter fluid rises while cooler, denser fluid sinks, driving vertical flow that redistributes heat and alters cooling rates, key factors in solidification and grain evolution [38]. Unlike Marangoni convection, which is driven by surface tension gradients along the melt pool surface (primarily horizontal), natural convection acts throughout the melt pool volume and is predominantly vertical [38]. Although generally weaker than Marangoni flow, natural convection has been shown to influence melt pool symmetry and grain orientation, especially in larger pools (e.g., DED). It is often modeled using the Boussinesq approximation and studied via CFD simulations [18]. Experimentally, its effect is inferred from asymmetries in melt pool geometry or grain structures (e.g., elongated columnar grains near the melt pool center) as shown in the bottom Fig. 1. For example, Liu et al. [39] used a multiscale thermal and PF model in electron beam powder bed fusion (EB-PBF) of Ti-6Al-4V to show how the interaction between thermal gradients and preferred growth directions controls grain orientation and texture. Likewise, Liu et al. [40] applied a 3D thermal-fluid and CA model in L-DED to highlight how melt pool convection, driven by Marangoni and buoyancy forces, shapes grain morphology.

**Evaporation and recoil pressure:** Recoil pressure, generated by intense laser-induced surface evaporation, plays a pivotal role in melt pool dynamics, particularly in LPBF and to some extent in DED, as shown in Fig. 2. The vapor jet exerts a downward force on the melt pool, creating a surface depression that enhances laser absorption through multiple reflections, potentially leading to keyhole formation at high energy densities [41]. Excessive recoil pressure can destabilize the melt pool, promoting spatter, denudation, and powder entrainment [42]. Its magnitude scales with laser power and scan speed and is highly sensitive to beam shape and intensity distribution [43]. Processing under vacuum or low ambient pressure, lowers the boiling point and amplifies recoil-induced vapor plume effects, while elevated pressures may suppress vaporization but increase plasma formation, which can further influence melt pool stability [44].

A deeper understanding of internal melt pool pressures has been advanced through computational modeling. Dai et al. [42] used a multiphase computational fluid dynamics (CFD) model to simulate the effects of recoil pressure on melt pool behavior in LPBF. Their work showed that recoil pressure strongly governs melt pool depression and spatter formation, which are critical for pore generation and microstructure evolution. Complementary studies also explored how vapor-induced recoil pressure contributes to instability in the melt pool, influencing solidification fronts and leading to defects such as keyholes and porosity [45]. Furthermore, they illustrated that recoil pressure gradients drive Marangoni and convective flows, which impact melt pool shape, solidification fronts, and dendrite orientation.

**Remelting:** Remelting in metal AM, whether intentional (e.g., via a second laser pass) or unintentional due to heat accumulation, plays a critical role in modifying microstructure by altering thermal history, temperature gradients, and cooling rates, especially in LPBF and DED processes [46,47]. The extent and effects of remelting are governed by parameters like laser power, scan speed, layer thickness, and overlap, and can be tailored using pulsed wave lasers [46–48]. This secondary thermal cycle influences grain structure variably, some studies report grain refinement through nucleation or fragmentation of columnar grains [47,49], while others observe grain coarsening due to prolonged high temperatures [50]. It also affects crystallographic texture, occasionally producing more randomly oriented grains that reduce anisotropy [47]. Moreover, remelting modifies solid-state phase transformations and precipitation behavior, affecting phase fractions, nanoprecipitate formation, e.g., Al$_3$(Sc,Zr), and ultimately, mechanical properties like microhardness, strength, elongation, and isotropy [47,51,52]. Additionally, the influence of in-situ laser polishing on pore defects is also observed. Collectively, these findings position remelting as a powerful strategy for microstructure and property tailoring in AM.

## 3. EXPERIMENTAL CHARACTERIZATION METHODS

Experimental characterization provides the essential physical foundation for understanding microstructure characteristics and evolution in metal AM processes like LPBF and DED [53]. Establishing clear relationships of process-microstructure-property is critical for validating both physics-based and data-driven predictive models [54–56]. While simulations offer valuable insights, their fidelity hinges on calibration and verification against the experimental data [8].

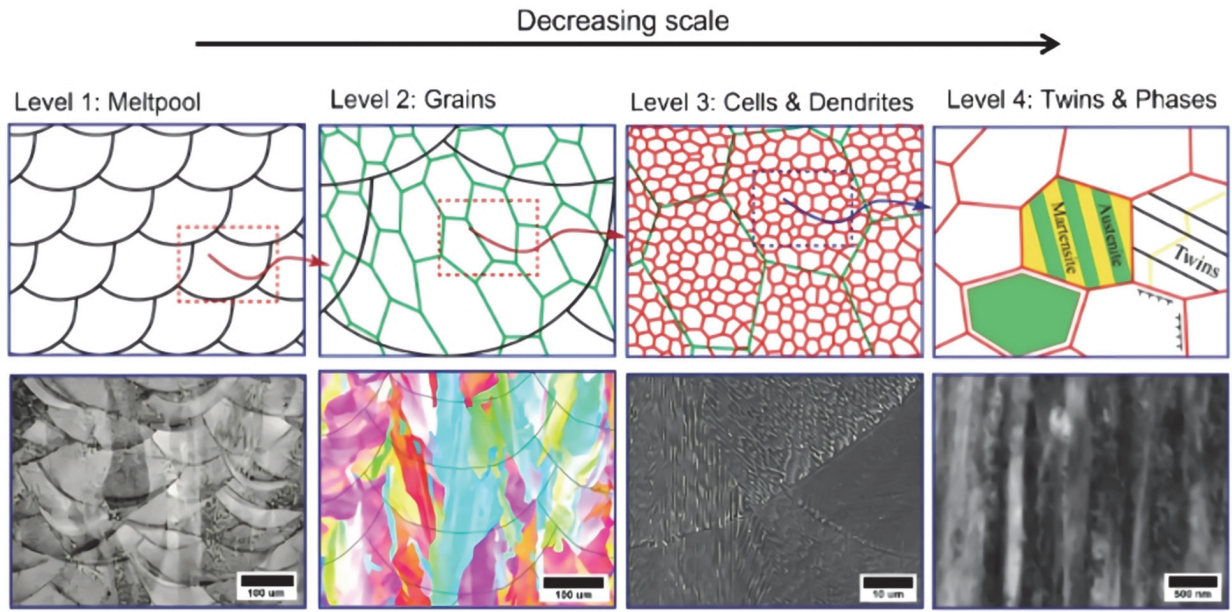

**FIG. 3:** Four microstructure levels in LPBF-processed SS316L: melt pool, grains, cells & dendrites, and twins & phases, shown in decreasing scale [56].

A wide range of experimental methods (see Table 1) is used to observe microstructures and defects at different length scales, as shown in Fig. 3. Techniques like Optical Microscopy (OM) and Scanning Electron Microscopy (SEM) are employed for basic microstructural analysis, while advanced tools such as Electron Backscatter Diffraction (EBSD), X-ray Computed Tomography (XCT), and Transmission Electron Microscopy (TEM) provide high-resolution crystallographic, 3D, or nanoscale insights. These methods not only support direct investigations but also serve as ground truth for refining simulations and training ML models [8]. Together, these methods enhance understanding of the complex process–microstructure–property relationships in metal AM. Their integration with computational approaches continues to drive advancements in AM research and industrial practice. A more detailed discussion of the experimental limitations and emerging strategies to address them is presented later in the review.

**Table 1:** Characterization Techniques and Applications in Metal AM

| Technique | Process | Application Focus | Ref. |
|---|---|---|---|
| OM | LPBF | Grain size, melt pool size, porosity | 57 |
| SEM | LPBF, DED | Melt pool, porosity (LOF/gas/keyhole), dendrites structure, surface analysis | 46,58 |
| EBSD | LPBF, DED | Grain orientation, size, texture, misorientation, β-grain | 39,59 |
| X-ray Diffraction (XRD) | LPBF, DED | Phases (α, β, martensite), texture | 60,61 |
| TEM | LPBF, DED | Dislocations, precipitates, substructures, strengthening | 62,63 |
| Energy Dispersive X-ray Spectroscopy (EDS/EDX) | LPBF, DED | Composition, phase ID, vaporization defects | 28,64 |
| XCT | LPBF, DED | 3D pore imaging: morphology, volume, distribution | 56 |
| In-situ Process Monitoring | LPBF, DED | Melt pool size, keyhole, spatter, balling (IR/Vis/Acoustic) | 65,66 |

## 4. PHYSICS-BASED COMPUTATIONAL MODELING

Physics-based approaches like PF, CA, and KMC simulate microstructure evolution by solving governing equations for nucleation, grain growth, and phase transformations. Physics-based computational methods are essential tools for understanding and predicting microstructure evolution during metal AM processes. These methods use fundamental physical principles to simulate the complex phenomena occurring during AM, providing insights into the process-microstructure-property relationships. The following subsections discuss different numerical techniques used to determine the microstructure in metal AM.

### 4.1 Phase-Field (PF) Method

PF modeling has emerged as a powerful mesoscale approach for simulating solidification phenomena in metal AM, particularly under the rapid cooling and complex morphologies encountered in LPBF and DED [67–69]. Unlike sharp interface models, PF employs thermodynamically consistent field variables (order parameters) governed by time-dependent Ginzburg-Landau (TDGL) equations [70], allowing implicit tracking of solid-liquid interfaces and grain boundaries [68,71]. PF models simulate nucleation, often based on undercooling, followed by competitive grain growth, dendritic evolution, and coarsening [72,73]. Its ability to resolve dendritic microstructures, such as primary dendrite arm spacing (PDAS), and predict features like microsegregation and crystallographic texture, makes PF especially suited for AM [74]. Studies by Radhakrishnan et al. [74] and Ma et al. [75] have demonstrated PF's strength in capturing columnar-to-equiaxed transition (CET), grain size distribution, and branching behavior under AM thermal conditions. Solute diffusion is inherently modeled, enabling prediction of microsegregation and phase formation, critical for local mechanical properties.

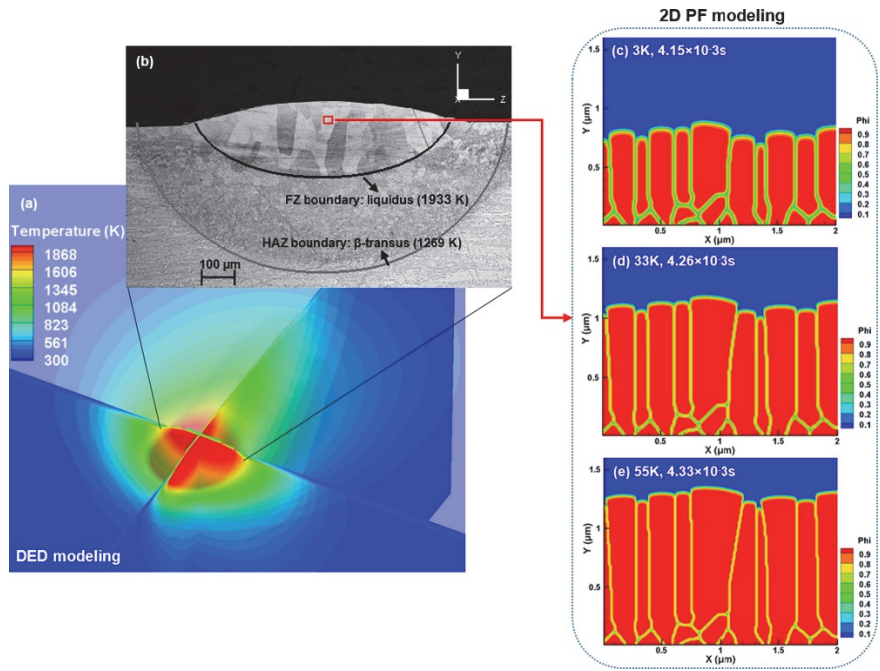

**FIG. 4:** Phase-field modeling framework for the solidification of Ti-6Al-4V, illustrating beta grain growth [67]. (FZ: Fusion Zone Boundary; HAZ: Heat Affected Zone Boundary).

PF also allows crystallographic anisotropy to be incorporated in energy and mobility terms, enabling prediction of texture aligned with directional heat flow, as often observed experimentally in AM builds [76,77]. While PF simulations are limited to mesoscale volumes (10–100s of μm) [68,73], as shown in Fig. 4, they provide indirect insight into defects like hot cracking via microsegregation patterns and can be coupled with stress models for further prediction [76,78]. Multi-physics integrations have enhanced PF's predictive capability, thermal fields from CFD or Finite Element Method (FEM) models [69,75,79] and melt pool dynamics from Lattice Boltzmann Methods [80] have been used to drive PF evolution more accurately. These developments and their applications are summarized in Table 2. Computational advances, including adaptive meshing, parallelization, and integration with ML, now enable large-scale 3D simulations of polycrystalline microstructures over microsecond to millisecond timescales. Compared to CA or KMC, PF's diffuse interface and thermodynamic rigor (often using CALPHAD databases for alloys like Inconel 718 and Ti-6Al-4V) make it a more fundamental and flexible tool for AM microstructure modeling [60,68,69,71,72]

**Table 2:** PF Modeling Applications in AM

| Process | Material | Application Focus | Ref. |
|---|---|---|---|
| LPBF | Ti-6Al-4V | Solidification (β grain) and β→α/α' transformation using hybrid PF + ML + FDMC | 60 |
| DED | H13 steel | Melting, solidification, dendrite growth, solid phase transformation, CALPHAD-based | 81 |
| LPBF | Ti-based alloys (Ti-45Al) | PDAS prediction, site-specific segregation, melt pool validation | 77 |
| LPBF | IN 718 | Solidification and multiphase modeling, integration with post-build heat treatment | 75 |
| LPBF/DED | Ni-based superalloys | Microstructure evolution, PF-FE-CALPHAD coupling | 74 |

### 4.2 Cellular Automata (CA) Method

CA modeling has become a widely adopted mesoscale approach for simulating microstructure evolution in metal AM [82], offering a strong balance between computational efficiency and physical realism [83]. CA discretizes the domain into a lattice of cells [10], where each cell updates its state (e.g., phase, grain orientation) based on local interaction rules and transient thermal fields, as shown in Fig. 5, typically sourced from FEM or Finite Volume Method (FVM) simulations [84]. This approach effectively captures key solidification mechanisms such as nucleation, anisotropic grain growth, competitive growth, and epitaxial extension across layers [85]. CA models operate over micrometer-to-millimeter spatial scales and microsecond-to-second timeframes [86], making them suitable for simulating representative AM volumes, including multiple melt tracks and layers [87]. Grain-scale phenomena, like fusion boundary nucleation, anisotropic dendritic growth, and competitive texture evolution, are well captured in 3D CA [7]. Grain morphologies such as columnar, equiaxed, and zigzag structures have been simulated for alloys like Inconel 718 and AA-2024, with average grain size predictions often within 10–15% of experimental data [88]. An overview of these advancements is given in Table 3. Compared to PF models, CA models are more computationally scalable while retaining strong physical representations of grain growth, texture evolution, and morphology formation in AM processes [10,54,83].

**Table 3:** CA Modeling Applications in AM

| Process | Material | Application Focus | Ref. |
|---|---|---|---|
| LPBF | 316L SS | Grain structure formation during solidification, using CAFE (CA + thermal FE) | 84 |
| DED | Ti-6Al-4V | Thermal history, grain morphology (dendritic, columnar), nucleation, growth orientation (CA-FE) | 89 |
| DED | Binary β-Ti alloy systems | Grain nucleation and growth, Columnar-to-Equiaxed Transition (CET), parameter effects | 83 |
| LPBF | Ti-6Al-4V, | Grain structure evolution (2D & 3D CAFD), experimental validation | 7 |
| LPBF | NiTi SMA | Microstructure evolution using numerical CA modeling approach | 90 |

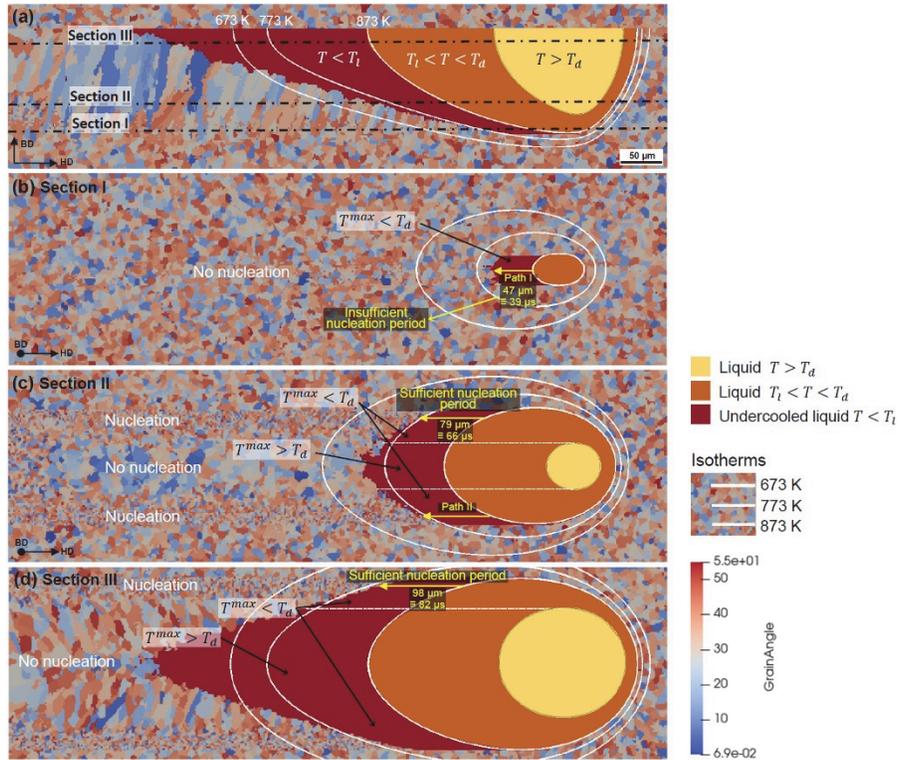

**FIG. 5:** Cellular Automata model illustrating grain boundary nucleation in LPBF of Al-alloy; (a) Transverse Direction (TD) section at mid-width and (b)–(d) corresponding Build Direction (BD) sections [85].

## 4.3 Kinetic Monte Carlo (KMC) Method

KMC modeling has become a key approach for simulating solidification phenomena in metal AM, particularly for predicting microstructure evolution at the mesoscale capturing [69]. These methods simulate the probabilistic evolution of microstructures through discrete events, such as atomic rearrangements and grain boundary migrations, driven by thermodynamic principles like the minimization of interfacial energy [50,91]. These models are well-suited to the stochastic nature of grain growth and texture development that occurs during AM solidification. Applications of KMC, such as the open-source SPPARKS simulator [92], exemplify its power in generating representative microstructures for AM processes.

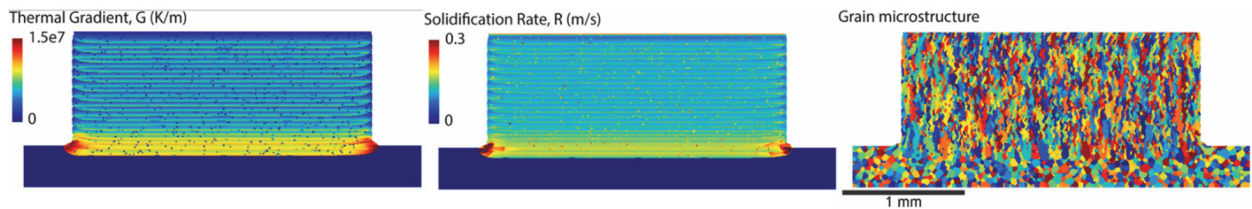

**FIG. 6:** KMC-simulated thermal gradient (G), solidification rate (R), and microstructure evolution for a thin-wall build, highlighting layer variations and nucleation effects [50].

A primary mechanism modeled in KMC is curvature-driven grain boundary migration, with events like atom attachment/detachment at solid-liquid interfaces, as shown in Fig. 6. Such models have been employed to study the effects of process parameters such as laser speed on grain morphology [93] and transient thermal fields on grain structure dynamics [94]. Furthermore, KMC

Page **12** of 36

methods predict crystallographic texture by incorporating orientation-dependent event rates, as demonstrated by Pauza et al. [95] and Whitney et al. [60], where coupled models like Finite Difference-Monte Carlo (FD-MC) help simulate texture evolution under complex AM heating cycles. Recent advancements have focused on improving the computational efficiency of KMC simulations using high-performance computing, as well as incorporating more realistic physical models, particularly for nucleation and grain competition [10,91]. A summary of these developments and their applications is presented in Table 4. In addition, KMC simulations have been integrated with data-driven approaches to better understand process-microstructure relationships [92]. Despite these advances, limitations in handling complex events like nucleation and grain growth under varied thermal conditions remain, highlighting areas for future improvement [75].

**Table 4:** KMC Modeling Applications in AM

| Process | Material | Application Focus | Ref. |
|---|---|---|---|
| LPBF | High-Mn steel | KMC-based 3D microstructure dataset generation; scan strategy effects | 92 |
| LPBF/DED | IN 625, 304L SS | Dynamic KMC for grain morphology prediction under evolving melt pool and HAZ conditions | 94 |
| LPBF | Ni-based superalloy | Texture-aware Potts model; effects of hatch spacing and layer thickness | 95 |
| DED | Ti–6Al–4V | Grain growth modeling in HAZ; solid-state phase evolution during thermal cycles | 96 |

## 5. DATA-DRIVEN MODELS

Data-driven techniques are capable of handling complex problems across various fields, including computer science, robotics, aviation, biomedical science, materials science, and manufacturing [97]. The inherent complexities and limitations of experimental approaches in metal AM have necessitated the adoption of data-driven modeling techniques, such as artificial intelligence (AI), particularly ML [6]. Specifically, researchers have applied ML techniques in different aspects of the metal AM, such as process control, design for AM, process monitoring, process parameter optimization, quality control, defect detection, and property prediction [97]. While experimental methods play a crucial role in characterizing AM processes, they often fall short in capturing the dynamic, fine-scale phenomena that occur during material deposition and solidification [6]. Moreover, the vast number of process parameters and their intricate interdependencies pose significant challenges for comprehensive exploration through traditional experimentation alone [98]. High computational costs of physics-based simulation models often limit their feasibility for real-time applications [14]. Given the high-dimensionality and non-linearity of AM processes, data-driven modeling has emerged as a powerful tool to complement physics-based models [98]. By leveraging experimental and simulation data, AI/ML techniques can improve accuracy, efficiency, and scalability in predictive modeling, ultimately accelerating advancements in AM technology [66]. This section provides a comprehensive review of the current state-of-the-art in data science-based modeling for microstructure prediction in AM, encompassing the diverse ML algorithms employed, the types of data utilized, and the key applications in prominent AM processes.

## 5.1 Machine Learning Models

A significant amount of work has focused on utilizing ML algorithms trained on experimental data to predict microstructural features in AM components. These approaches aim to learn the intricate, often non-linear relationships between input variables (e.g., process parameters, in-situ monitoring data) and output microstructural characteristics (e.g., grain size, phase distribution, defect formation). A wide array of ML algorithms has been employed, encompassing supervised learning, unsupervised learning, and deep learning paradigms [99]. Supervised learning algorithms, which learn from labeled datasets, are particularly dominant and include linear models (Linear Regression, Logistic Regression), instance-based learning (K-Nearest Neighbors - KNN), tree-based models (Decision Trees, Random Forests, Gradient Boosting), Support Vector Machines (SVM), Neural Networks (fully connected NN, convolution NN, Recurrent Neural Networks (RNNs)), Long Short-Term Memory networks (LSTMs), and Gaussian Process Regression (GPR). Unsupervised learning techniques such as clustering (K-means, Hierarchical clustering, Gaussian Mixture Models) and dimensionality reduction (Principal Component Analysis - PCA) are often used for exploratory data analysis and feature engineering.

Deep learning (DL), a subset of ML employing multi-layered neural networks, has gained substantial traction, with CNNs, RNNs/LSTMs, Generative Adversarial Networks (GANs), and Variational Autoencoders (VAEs) being prominent architectures [100]. CNNs excel at handling image-based representations of microstructures, commonly derived from experiments or simulations, by learning hierarchical spatial features directly from raw pixel data through convolutional and pooling layers. Architectures like U-Net and 3D CNNs have proven effective for tasks such as classification, semantic segmentation, object detection, and direct property prediction [59]. These models also act as powerful feature extractors, enabling dimensionality reduction and integration with other ML models like RNNs or LSTMs for modeling temporal evolution [101,102]. Despite their strengths, CNNs treat microstructures as regular grids, which may limit their ability to capture irregular grain interactions, an area where GNNs, with their topological awareness, offer a complementary advantage [103].

Graph Neural Networks (GNNs) emerge as a powerful tool for predicting the properties of polycrystalline materials by leveraging the graph-like structure of microstructures generated through simulations. GNNs provide a robust approach for analyzing the complex topology of polycrystalline microstructures in metal AM [104]. By representing individual grains as nodes and their boundaries as edges, GNNs effectively capture hierarchical relationships critical to determining material behavior. This graph-based representation allows GNNs to inherently understand grain adjacency and connectivity, fundamental aspects of microstructural evolution. Unlike CNNs, which treat microstructure images as regular grids, GNNs operate directly on the irregular structure of grains, learning features that remain invariant to spatial arrangement and grain count [103,104]. Through message-passing mechanisms, GNNs aggregate information from neighboring grains, enabling the network to capture the influence of local microstructural environments on both individual grains and overall material behavior. This capability positions GNNs as a potentially superior alternative to CNNs in microstructure analysis tasks. For example, Dai et al. [105] employed GNNs for predicting magnetostriction, while Thomas et al. [103] applied them to fatigue damage prediction. Several GNN architectures have been applied to microstructure analysis in AM. Graph Convolutional Networks (GCNs), which generalize convolution to graph data, have been utilized for feature extraction based on local grain neighborhoods [106]. Although Graph Isomorphism Networks (GINs) have not been widely explored in AM microstructure analysis, their high expressive power suggests they hold potential for future applications [103].

Additionally, Physics-Embedded Graph Networks (PEGN), proposed by Xue et al.[71], as shown in Fig. 7, reformulate the phase-field problem of microstructure evolution in AM as an unsupervised machine learning task on a graph, significantly accelerating simulations.

Additionally, RNNs and LSTM networks, like GrainNN[101] and GrainGNN[104], have been used to model the temporal evolution of microstructures under rapid solidification. These models, including PEGN and GrainGNN, not only enhance microstructure representation but also predict material properties, serving as efficient surrogate models to reduce the need for extensive experimentation. Their ability to model the evolution of grain structures under varying processing conditions is essential for designing materials with tailored properties.

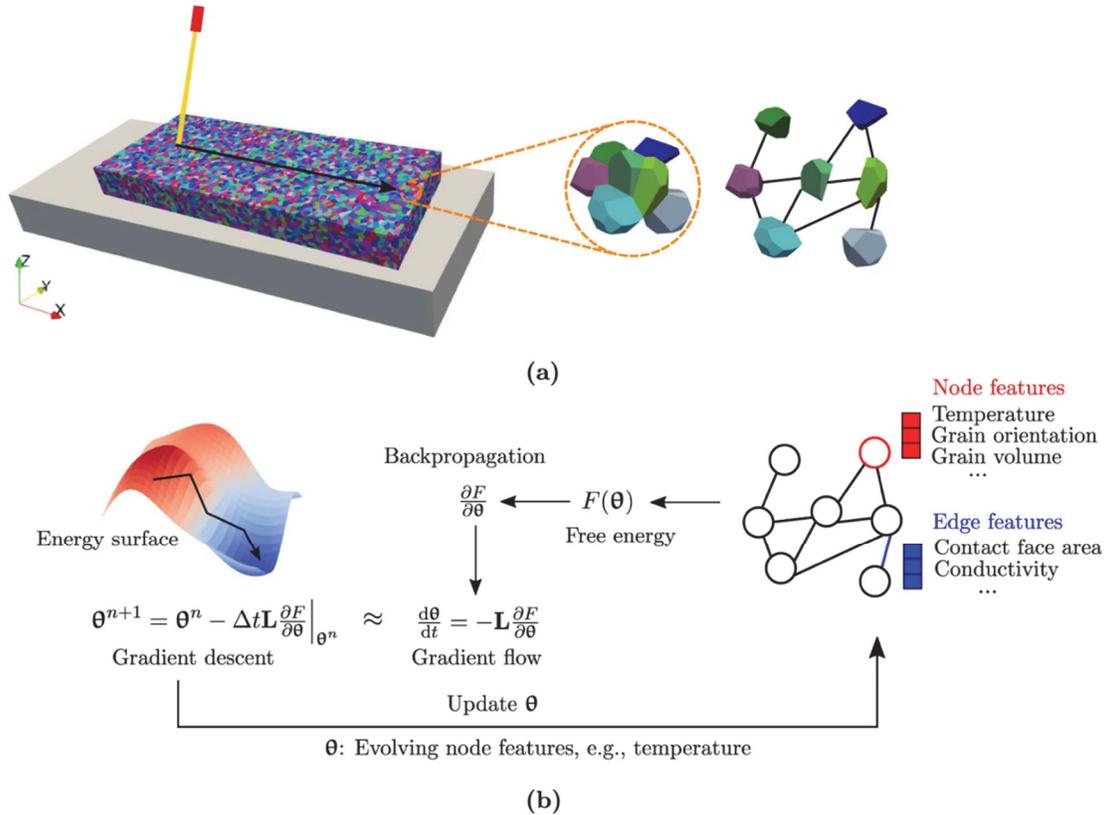

**FIG. 7:** PEGN framework for single-track PBF, modeling grain evolution as a graph with iterative feature updates via gradient descent[71].

### 5.1.1 Features of ML models

ML models used for microstructure prediction in metal AM typically rely on diverse input features tailored to the modeling objective. These include process parameters (e.g., laser power, scan speed), thermal history, grain orientation, microstructural images, in-situ monitoring data (e.g., thermal signatures, AE signals), and material-specific attributes like chemical composition or powder characteristics. The outputs range from grain structure features, such as morphology, size, and orientation, to defect classification (e.g., pores, voids) and mechanical properties (e.g., yield strength, ultimate tensile strength).

Recent studies showcase the variety of ML methods adapted to LPBF and DED processes. For instance, CNNs, including 3D CNNs and U-Net architectures, as shown in Fig. 8, have been applied for grain structure and defect prediction[68,107]. Generative models like cGANs are used for reconstructing microstructural features from processing parameters, while LSTM models such as GrainNN predict time-resolved grain growth. Other approaches include artificial neural networks (ANNs) that estimate grain growth from thermal gradients, and hybrid CNN-wavelet models to infer mechanical property distributions. These techniques demonstrate the growing application of ML in modeling process-microstructure-property relationships. A summary of ML-based studies for microstructure modeling used in AM is provided in Table 5.

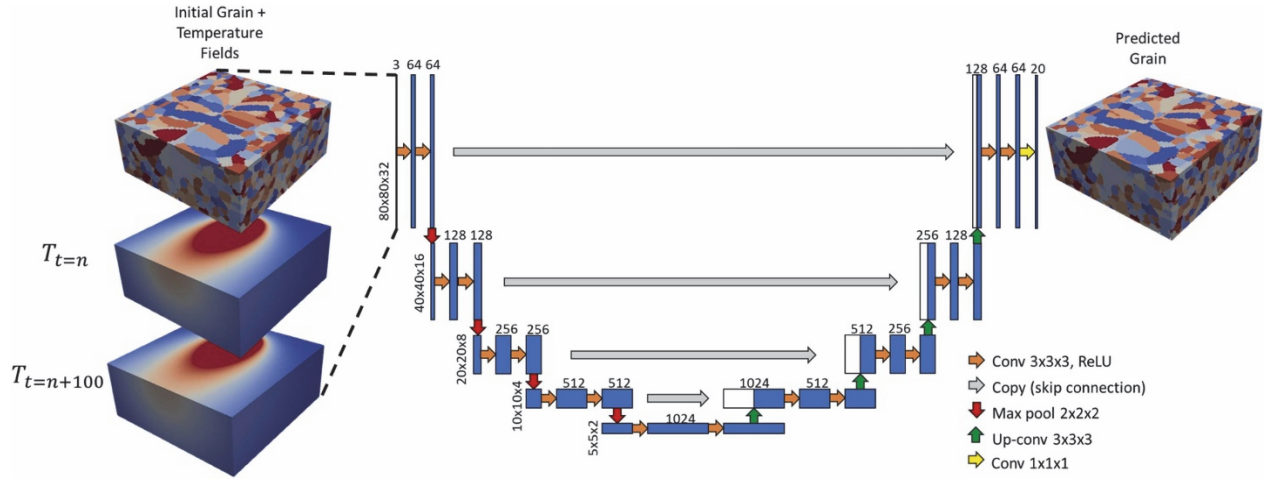

**FIG. 8:** 3D U-Net for microstructure prediction using grain orientation and temperature inputs with coarse-grained time steps [68].

**Table 5:** ML Models for Microstructure and Property Prediction in Metal AM

| Process | ML Model | Input | Output | Ref. |
|---|---|---|---|---|
| LPBF | VGGNet (3D CNN) | Grain ID, crystal orientation | Mechanical properties | 107 |
| LPBF | Modified U-Net (CNN) | Microstructure images | Elastic stress fields, defect mapping | 58 |
| DED | CNN + Wavelet Transform | Thermal history | Ultimate tensile strength | 108 |
| LPBF | Conditional GAN (cGAN) | Laser power & speed | Microstructural feature prediction | 109 |
| DED | ANN | Thermal gradient, crystal orientation | Grain growth prediction | 110 |
| LPBF | LSTM (GrainNN) | Temporal grain growth data | Epitaxial grain growth prediction | 101 |

### 5.1.2 Training data

ML models for microstructure prediction in metal AM are trained using both experimental and synthetic data. Experimental data typically includes images measured by SEM, EBSD, OM, and XCT, along with process parameters like laser power, scan speed, and build orientation. These datasets serve as inputs for models such as CNNs, which are well-suited for analyzing microstructural features. CNNs are widely used for tasks including defect detection, grain size estimation, porosity analysis, and phase segmentation. GANs, particularly cGANs, further enhance experimental datasets by generating synthetic microstructures, helping to address issues of data scarcity and cost. These ML applications using experimental data are summarized in Table 6.

Table 6: ML Applications Using Experimental Data in Metal AM

| Experimental Data | Process | ML Model | Application | Ref. |
|---|---|---|---|---|
| Micrograph Images (OM, SEM) | LPBF | cGAN | Predicted α-phase / martensite morphology and size | 109 |
| Pyrometer & High-speed Camera | LPBF | SeDANN | Predicted the melt pool width | 111 |
| Thermal & Optical Tomography Images | LPBF | KNN | Predicted porosity, melt pool depth, and grain size | 13 |
| High-speed Imaging | LPBF | SVM, MLP, KNN, RF, CNN | Detected various defects (keyhole, under-melting, balling) | 58 |

To broaden predictive capabilities, simulation-generated data from physics-based models such as PF, CA, and KMC modeling are also utilized. These simulations replicate microstructure evolution under varied processing conditions, offering extensive and controllable datasets for training. ML models trained on such synthetic data can be used for surrogate modeling, parameter calibration, and accelerated simulations, providing insights into scenarios that are difficult to capture experimentally. The integration of both experimental and simulation data has significantly improved microstructure prediction across AM processes like LPBF, DED, and EPBF. A summary of ML-integrated simulation frameworks used in AM is provided in Table 7.

Table 7: ML–Integrated Simulation Frameworks for Microstructure Prediction in Metal AM

| Simulation | Process | ML Model | Applications | Ref. |
|---|---|---|---|---|
| PF Method | LPBF | CNNs / 3D U-Net | 3D grain structure prediction, surrogate modeling for complex PF simulations | 68 |
| | LPBF | Diffusion Probabilistic Field Model | Captures irregular and realistic grain morphologies for microstructure generation | 112 |
| CA Method | DED | Neural Networks (NN) | CA-FVM + NN to predict grain shape/aspect ratio from thermal history | 113 |
| KMC Method | LPBF/DED | LSTM-SE | Surrogate time-dependent modeling of precipitate kinetics | 114 |

### 5.1.3 Hybrid Machine Learning Approaches

To enhance predictive capabilities and better capture the complex spatio-temporal evolution of microstructures in metal AM, hybrid ML models are gaining traction. A particularly promising direction involves combining CNNs with RNNs or LSTM architectures. In these models, CNNs extract spatial features from microstructure images, while RNNs or LSTMs model their temporal evolution, enabling predictions of grain growth or phase transformation over time [115]. Additionally, using pre-trained CNNs as feature extractors followed by traditional ML regressors can be advantageous when data is scarce. This transfer of learned spatial features minimizes the need for extensive domain-specific training [100].

To bridge simulation and experimental domains more effectively, advanced hybrid learning strategies, such as hierarchical networks, multi-task learning, and transfer learning, are also gaining adoption. These approaches reduce model complexity, enhance generalization, and embed physical fidelity. Representative examples of these methods and their implementation strategies are summarized in Table 8.

**Table 8:** Hybrid ML Methods for Microstructure Prediction in Metal AM

| Hybrid Method | Input | Output | Remarks | Ref. |
|---|---|---|---|---|
| CNN + RNN / LSTM | Microstructure images | Grain growth, phase evolution | Learns spatial features and their temporal dynamics | 115 |
| Hierarchical Networks | Thermal history | Microscale grain morphology, phase fractions | ML surrogates couple mesoscale and microscale predictions | 113 |
| Multi-task Learning | Thermal history, process parameters | Grain size, orientation, phase fraction | A single model simultaneously predicts multiple microstructural features | 98 |
| Transfer Learning | Simulation-trained CNN features | Experimental microstructure predictions | Enhances real-world predictions using limited experimental data | 100 |
| Physics-Data Fusion | Simulation data, in-situ/optical observations | Microstructure evolution (Potential) | Combines physics-based modeling with data-driven learning | 14 |

These hybrid approaches are instrumental in advancing predictive modeling for metal AM. Expanding on such methodologies, hierarchical multi-scale models have combined FD-MC simulations for part-scale thermal prediction with PF modeling of α/α' transformations, where ML surrogates effectively bridge scale transitions [60]. Similarly, multi-task learning frameworks have demonstrated the ability to simultaneously predict grain morphology and phase fractions from processing parameters and thermal histories, enhancing generalization in data-sparse settings. Transfer learning further strengthens this hybrid toolkit, by leveraging simulation-trained models and fine-tuning with sparse experimental data, it significantly boosts predictive performance for real-world applications. Furthermore, Ren et al. [116] combined high-speed synchrotron X-ray imaging, in situ thermal imaging, and physics-based simulations to investigate pore defect formation caused by unstable keyhole oscillations in LPBF of Ti-6Al-4V, while Yadav et al. [117] integrated data-driven and physics-driven components to model grain evolution in laser DED. Likewise, PINNs offer a compelling framework by embedding physical constraints directly into learning, especially powerful when experimental data is limited but physical laws are well established.

## 5.2 Physics-Informed Machine Learning (PIML)

PIML has emerged as a promising approach for modeling AM processes, especially in data-scare environment or when dealing with the noisy data, offering a potential solution to the limitations of traditional experimental, analytical, and numerical methods [118,119]. These traditional methods often suffer from drawbacks such as extensive setup times, high costs, difficulties in generalization, and significant computational burdens [120]. PIML aims to address these challenges by fusing physical laws with neural networks, enabling the development of process-microstructure-property models that are rapid, generalizable, scalable, and transferable across diverse AM processes, machines, and materials, while maintaining accuracy under varying processing conditions [11,121–123].

A core component of PIML is the incorporation of physical laws, expressed as partial differential equations (PDEs) (e.g., conservation of mass, energy and momentum, heat transfer equations), into the training process. This is achieved by adding loss terms to the loss function that penalizes deviations from the governing equations [119]. This embedding of physics enhances interpretability, reduces data requirements, and ensures physically plausible predictions. Figure 9 illustrates a typical PINN architecture applied to metal AM, showcasing the integration of physical laws into the training process. A baseline PINN's loss function can be expressed as:

$$\mathcal{L} = \mathcal{L}_{PDE} + \mathcal{L}_{BC} + \mathcal{L}_{IC}$$

where these terms represent the losses associated with the PDE, boundary conditions, and initial conditions, respectively [119,124].

PINNs approximate a mesh-free approach [125,126], solutions to governing conservation equations (e.g., PDEs), melt pool dynamics, grain nucleation and growth dynamics by minimizing a loss function that integrates physics-based constraints with initial/boundary conditions, and also some experimental or simulated data [122,123]. This allows for incorporating prior knowledge into the model [127]. Modern deep learning libraries have enabled the use of automatic differentiation (AD), which ensures efficient computation of gradients, reducing the computational cost associated with traditional numerical simulations [126]. While AD is commonly used, the possibility of replacing it with an approximation of a differentiation operator to further decrease computation time and potentially guarantee a convergent rate has been explored [128]. Thus, in principle, PINNs can be completely unsupervised when the data is not readily available and boundary or initial conditions are well defined [119].

A range of neural network architectures has been proposed to enhance the predictive power of PINNs in AM. CNNs, LSTMs, and GNNs have been integrated into PINN frameworks, each contributing to specific modeling challenges [124]. CNNs have shown effectiveness in image-based microstructure prediction [121], while LSTMs excel in modeling time-dependent solidification processes [127]. GNNs, in particular, have advanced the accuracy of grain structure predictions by efficiently handling irregular, non-Euclidean geometries, which are common in AM [71,124].

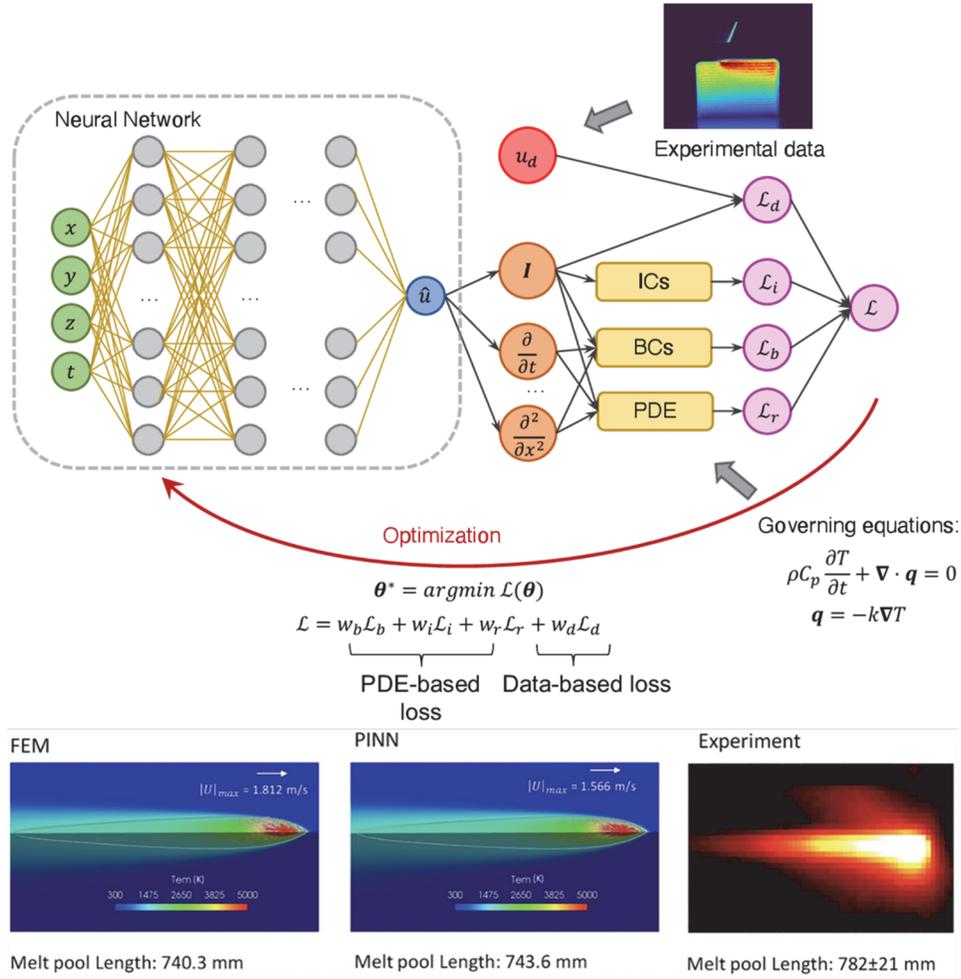

**FIG. 9:** PINN framework for LPBF integrating real-time thermal data and physics-based constraints (top) [118], and comparison of melt pool predictions from FEM, PINN, and experiments (bottom) [120].

**5.3 Applications of PINN-based Microstructure Modeling in AM**

Unlike conventional DL approaches, which rely heavily on large-labeled datasets, PINNs embed fundamental physics laws, such as conservation of mass, energy and momentum, heat transfer equations, and grain growth dynamics, into their learning process. This allows them to predict temperature fields [122,123,125,129], melt pool dynamics [71,128], porosity formation [11,130] and thermal stresses [131]. With this background, PINNs has potential application in predicting microstructural features like grain morphology, phase distributions, and defect formations under varying processing conditions.

The ability of PINNs to handle multi-scale phenomena is particularly beneficial for metal AM, where processes span a wide range of scales. In AM, accurately capturing the transient temperature fields is essential, as these thermal gradients drive grain growth and the overall evolution of microstructures. PINN models can be tailored to focus on different scales by adjusting their loss functions to emphasize fine details in temperature gradients, even within smaller melt pools. As shown in Table 9, PINNs have demonstrated effectiveness in predicting temperature fields during metal AM, with accuracy levels varying across different AM processes. For instance, recent studies have demonstrated that transfer learning-based PINN models can predict the 3D temperature distribution during single-track metal deposition with high accuracy, with average temperature prediction errors reported to be below 1.3 % [125], despite the limited availability of labeled data.

**Table 9:** PINN Applications for Thermal Prediction and Melt Pool Dynamics in AM

| Process | Method | Input | Output | Ref. |
|---|---|---|---|---|
| DED | Transfer Learning-based PINN | IR images, simulation data | 3D temperature field | 125 |
| DED | PINN | FEM simulation data, validated by DED | Temperature field | 132 |
| LPBF/DED | ANN-based PIML | Simulation data | Temperature, melt pool dimension | 133 |
| LPBF/DED | Thermoelastic PINN | Process parameters, FEM data | Temperature (thermoelastic behavior) | 134 |
| DED | PIML | No labeled data | 3D temperature field | 135 |
| LPBF/DED | RAA-PIML | Temperature – small labeled simulations | Temperature field, melt pool morphology | 122 |
| LPBF/DED | Two-level PIML | Process parameters, pre-scan temperature | Melt pool size | 136 |

In addition to temperature prediction, PINNs have proven effective in modeling melt pool dynamics. By incorporating temperature-dependent material properties [132] into customized loss functions based on physical laws [125,137], these models can accurately estimate both the size and temperature of the melt pool [129,134]. They are even capable of inferring key process parameters such as the Reynolds and Peclet numbers from temperature and velocity data using an inverse approach [131]. This capability extends to predicting complex behaviors like melt pool dynamics under the influence of Argon gas-driven shear flows, without using any training data on velocity and pressure, thereby avoiding the need to directly solve the notoriously challenging nonlinear Navier–Stokes equations [131].

The evolution of microstructures also depends on grain growth and solidification processes, which are intimately linked to the material's thermal history. Recent advancements include the integration of PINNs with other computational methods to model microstructure evolution, as shown in Table 10. For example, a study detailed the calibration of a thermo-microstructural model for LPBF of Hastelloy X by integrating a PINN for thermal analysis with a CA model for microstructure simulation [138]. This approach leveraged the computational efficiency of PINNs for thermal model calibration through inverse analysis based on experimental melt pool dimensions, ultimately identifying the optimal CA parameters to represent observed microstructures. Similarly, Liu et al. [139] also explored a hybrid physics-based data-driven process design framework using physics-constrained neural networks to construct surrogates of process-microstructure relationships for optimizing dendritic growth in alloys like Ti-6Al-4V. Kats et al. [113] proposed a PIML framework for refining DED process parameters to achieve specific grain microstructures by using high-fidelity numerical data and extracting thermal gradients and cooling rates to predict grain size and aspect ratio.

Table 10: PINN Applications for Microstructure Prediction in AM

| Process | Method | Input | Output | Ref. |
|---|---|---|---|---|
| LPBF/DED | PCNN with Bayesian Optimization | PF simulation with varying parameters | Dendritic area, micro segregation | 139 |
| DED | PIML | High-fidelity numerical data | Grain microstructures | 113 |
| LPBF | PINN for thermal & CA microstructure model | Experimental melt pool dimensions | Temperature, melt pool, microstructure | 138 |

### 5.4 PINNs: Current Status and Future Directions

The current active area of development involves modifications to the baseline PINN framework. Ongoing research includes the use of adaptive activation functions to improve learning performance and convergence in solving differential equations [140]. Efforts are also being made to implement domain decomposition and preconditioning strategies, which aim to improve scalability and accuracy when handling large spatio-temporal domains through parallel computing [141]. In addition, alternative sampling methods are being investigated to increase training efficiency in regions with sharp gradients or complex physical behavior [142].

A significant advancement in this field is the integration of PINNs with in-situ monitoring techniques, such as real-time thermal imaging (see Table 9). By incorporating real-time sensor data into the PINN framework, these models can dynamically update their predictions, leading to more accurate process control and optimization. This approach is further reinforced by the development of physics-informed online learning models, which continuously update their weights as new data becomes available, thereby improving real-time process control [123]. Another promising direction is Transfer Learning and Domain Adaptation, which enhances the generalization capabilities of PINNs for metal AM applications [125]. These methods involve pre-training PINNs with simulation data (see Table 9) and fine-tuning them with experimental data, reducing training time and improving prediction accuracy. This approach enables PINNs to leverage computationally generated data while ensuring robust performance when applied to real-world experimental observations. Given the inherent uncertainties in AM processes, researchers are actively developing probabilistic PINN models to enable uncertainty-aware predictions [143–146]. These models provide confidence intervals for predictions, offering a more comprehensive understanding of model reliability and robustness. Finally, ongoing research is focused on architectural enhancements for PINNs in AM applications [11,125]. This includes the incorporation of lightweight attention mechanisms, ResNet blocks, and fully connected layers to better capture spatiotemporal correlations within complex AM processes while maintaining computational efficiency. These architectural enhancements aim to boost the model's ability to learn and represent the intricate physical dynamics that govern microstructure evolution.

# 6. CHALLENGES AND OUTLOOK

## 6.1 Experimental characterization challenges

Despite advances in characterization techniques, significant challenges remain in capturing the microstructure evolution during metal AM. A major limitation is the inability to directly observe microstructural changes during the rapid melting and solidification phases. While in-situ monitoring of thermal fields and melt pool dynamics offers indirect insights, these methods do not visualize the evolving solid structure itself. Establishing clear relationships between the wide range of process parameters and the resulting hierarchical microstructure across multiple length scales is difficult and requires extensive experimentation. This effort is further hampered by the lack of high-throughput and standardized characterization methods, as well as difficulties in managing the large datasets produced by modern techniques.

Moreover, comprehensive experimental campaigns remain costly and time-consuming, acting as a bottleneck for validating simulation models. Similar to other approaches, the high cost of experiments under complex conditions, such as the unstable and highly dynamic nature of the melt pool with rapid thermal and flow fluctuations, along with its microscopic scale that complicates in-situ observation, limits the availability of validation data. Even simulation-guided methods, like the KMC approach, ultimately depend on rigorous experimental validation. Addressing these challenges requires innovative experimental methods. Emerging techniques such as synchrotron-based high-speed X-ray imaging and diffraction offer promising in-situ capabilities. Correlative microscopy, combining XCT, SEM/EBSD, and TEM on the same sample, provides a more comprehensive multi-scale perspective. Additionally, ML tools can accelerate the analysis of complex image and diffraction data, supporting the generation of high-fidelity datasets needed to validate both physics-based and data-driven models.

## 6.2 Numerical modeling challenges

Recent progress in numerical modeling has significantly advanced our understanding of microstructure evolution in metal AM, particularly in LPBF and DED. However, simulating the rapid solidification and non-equilibrium conditions intrinsic to AM remains computationally challenging. High-fidelity models such as PF are particularly resource-intensive, limiting their scalability for large-scale or multi-layer simulations, and thus driving ongoing efforts in algorithmic optimization. Although more efficient, CA and KMC methods still face challenges in modeling full-part geometries and often rely on empirical parameters that are difficult to calibrate under AM-specific conditions. Table 11 summarizes key computational approaches for microstructure modeling, outlining their features, strengths, and numerical limitations. Additionally, critical phenomena such as solute segregation, re-melting, and defect formation, key to grain morphology and texture, are either oversimplified or neglected in most models.

**Table 11:** Comparison of Modelling Approaches

| Approach | Key Features | Advantages | Limitations |
|---|---|---|---|
| PF | Continuous representation, multi-physics | High fidelity; detailed interface tracking | High computational cost; parameter sensitivity; Effective for small time scale |
| CA | Discrete, rule-based evolution | Fast, captures local interactions | Limited resolution; oversimplification |

| KMC | Probabilistic, event-driven simulation | Effective for large-time scale | Computationally intensive for large systems |
| Data-Driven | Uses large datasets; various architectures | Fast predictions; transferable to new data | "Black box" behavior; high data requirement |

Beyond individual modeling approaches, developing accurate multi-scale frameworks that couple different physical phenomena across length and time scales remains a core challenge. The hierarchical nature of AM solidification, from melt pool behavior to grain evolution and atomic-scale interactions, necessitates integrated models that can resolve process–microstructure–property linkages. Coupled frameworks that integrate CFD, FEM, and microstructure models show promise, but these are often computationally expensive and difficult to validate experimentally. Sequential or weakly coupled strategies, while more feasible, can suffer from data mapping inaccuracies across differing grids or solvers, ultimately reducing predictive accuracy. Bridging these gaps will require the adoption of hybrid approaches, particularly ML techniques, including PINNs, to enable faster and more accurate simulations across scales.

## 6.3 ML challenges

Despite recent progress, purely data-driven models for microstructure prediction in metal AM face key limitations. A major hurdle is the lack of comprehensive, high-quality, and well-labeled datasets linking process parameters to detailed 3D microstructural outcomes. The multi-scale, multi-physics nature of microstructure evolution, driven by rapid solidification, complex thermal gradients, and non-equilibrium effects, is difficult to fully capture with "black-box" ML models. While ML offers speed and scalability, these models often lack physical interpretability and may produce unreliable predictions, especially when extrapolating beyond the training domain or across different length scales. Integrating physics-based knowledge into ML improves generalization, but balancing accuracy, scalability, and fidelity across scales remains challenging. Furthermore, many existing models primarily focus on predicting static microstructural features, with limited capability to capture the dynamic evolution of microstructure during processing.

A key challenge is generalizability across different AM processes, as models trained on one process (e.g., PBF) may not perform well on others (e.g., DED) due to differences in thermal histories and grain evolution. This underscores the need for standardized data-sharing protocols and public benchmark datasets, such as AMMD and AM-Bench, which, similar to MNIST and ImageNet in computer vision, could significantly enhance ML transferability in AM. Additionally, DL models often suffer from high computational cost, lack of interpretability, and an inability to quantify uncertainty, which limits their reliability in decision-making contexts. However, challenges remain in modeling dynamic microstructure evolution, managing computational cost, and improving model explainability, making physics-informed and uncertainty-aware hybrid models essential for trustworthy and robust AM predictions. Recent advances in Uncertainty Quantification (UQ) have addressed this by distinguishing between aleatoric (process-induced) and epistemic (model-related) uncertainties. Techniques such as Bayesian Neural Networks (BNNs), probabilistic models and ensemble learning are gaining traction. Advancing uncertainty-aware, explainable ML frameworks is crucial for trustworthy and scalable predictive modeling in AM.

**6.4 PIML modeling outlook**

PIML has emerged as a promising framework for modeling microstructural evolution in metal AM. By embedding governing physical laws into neural networks, PINNs provide a mesh-free alternative to traditional PDE solvers, enabling simulation of complex phenomena such as heat transfer and grain growth. However, challenges persist in training stability, high computational cost, and generalization across materials, geometries, and process conditions. Ongoing research is exploring solutions such as domain decomposition, dynamic loss weighting, and adaptive sampling to improve training efficiency and accuracy. Transformer-based PINNs and neural operators are also being investigated to address issues like spectral bias, gradient imbalance, and ill-conditioning. Despite being designed for low-data regimes, PINNs still rely on high-quality experimental or simulated data, and errors such as "propagation failure" can accumulate without proper initialization or boundary conditioning.

When modeling highly complex and multi-physics phenomena like microstructural evolution, the underlying PDEs can be too intricate for PINNs to learn effectively without anchoring through experimental or numerical data. While PINNs offer flexibility across spatio-temporal scales, unlike traditional numerical methods constrained by time-stepping schemes, convergence can be slow or unstable under stiff or poorly scaled conditions. Without sufficient grounding, error accumulation and unreliable extrapolation can occur. Another limitation of PINNs lies in their application to multi-physics, high-dimensional AM processes such as LPBF, where balancing constraints or hyperparameter from different physical domains remains difficult. Addressing these challenges requires improved strategies for UQ, such as Bayesian PINNs, and enhanced data handling methods, including transfer learning and data augmentation. Recent advances have also focused on error control in noisy or sparse datasets and incorporating dynamic activation functions to better represent unknown conditions. To make PINNs viable for real-time adaptive control in AM, future work must prioritize scalable algorithms, robust training strategies, and high-performance computing integration. Interdisciplinary collaboration will be essential to bridge the gap between data-driven models, physics-based principles, and practical AM implementation.

**7. CONCLUSIONS**

This work has assessed the state-of-the-art of microstructure modeling for metal AM, with a particular focus on processes such as LPBF and DED. Foundational experimental techniques remain essential for characterizing microstructural features and validating predictive models. Traditional computational methods, such as PF, CA, and KMC, offer valuable mechanistic insights into phenomena like grain growth and solidification dynamics, but they face limitations in scalability, computational cost, and complexity. Meanwhile, data-driven approaches using ML have shown significant promise in capturing complex process–microstructure relationships. However, these models often require large, high-quality labeled datasets and high computation cost, and tend to operate as "blackbox", limiting interpretability and generalizability. These challenges emphasize the growing need for physics-informed learning frameworks that can enable accurate yet data-efficient microstructure prediction, process optimization, and consistent part quality in AM.

PIML, particularly PINNs, has emerged as a powerful paradigm that bridges the gap between physics-based modeling and data-driven modeling. By embedding governing physical laws into the training process of neural networks, PINNs offer improved generalizability, reduced data dependence, and enhanced interpretability. Applications of PINNs to thermal field prediction and melt pool dynamics have demonstrated great potential, offering mesh-free solutions and achieving high accuracy even with small, labeled data, and ongoing research is now extending their use to direct microstructure prediction. Future efforts may focus on improving training efficiency, incorporating uncertainty quantification, adopting advanced neural architectures such as CNNs, LSTMs networks, and GNNs, and integrating real-time in-situ sensing for adaptive process control. As PIML progresses, interdisciplinary collaboration and the convergence of experimental, computational, and AI-driven approaches will be essential for building robust, scalable, and intelligent AM systems capable of producing defect-free, high-performance components with tailored microstructures.

**ACKNOWLEDGMENT**

The authors would like to thank the financial support of the National Science Foundation under the grants CMMI-2323083 and CMMI-2412395.